\documentclass[10pt,twocolumn,letterpaper]{article}

\usepackage{cvpr}
\usepackage{times}
\usepackage{epsfig}
\usepackage{graphicx}
\usepackage{amsmath}
\usepackage{amssymb}

\usepackage{booktabs}       
\usepackage{url}
\usepackage{multirow}
\usepackage{wrapfig}
\usepackage{amsmath}
\usepackage{amsfonts}       
\usepackage{nicefrac}       
\usepackage[numbers,sort,compress]{natbib}

\DeclareMathOperator*{\argmax}{argmax}

\usepackage[pagebackref=true,breaklinks=true,letterpaper=true,colorlinks,bookmarks=false]{hyperref}

\cvprfinalcopy 
\def\cvprPaperID{44}

\ifcvprfinal\fi

\begin{document}

\title{Class-Incremental Learning with Generative Classifiers}

\author{Gido M. van de Ven$^{1,2,}$\thanks{Corresponding author: \tt ven@bcm.edu}~,~ Zhe Li$^1$ \& Andreas S. Tolias$^{1,3}$ \\ \vspace{-0.165in} \\
\normalsize$^1\!$ Center for Neuroscience and Artificial Intelligence, Baylor College of Medicine, Houston, Texas, USA \\
\normalsize$^2\!$ Computational and Biological Learning Lab, University of Cambridge, Cambridge, United Kingdom \\
\normalsize$^3\!$ Department of Electrical and Computer Engineering, Rice University, Houston, Texas, USA
}

\maketitle
\thispagestyle{plain}
\pagestyle{plain}

\begin{abstract}
Incrementally training deep neural networks to recognize new classes is a challenging problem. Most existing class-incremental learning methods store data or use generative replay, both of which have drawbacks, while `rehearsal-free' alternatives such as parameter regularization or bias-correction methods do not consistently achieve high performance. Here, we put forward a new strategy for class-incremental learning: generative classification. Rather than directly learning the conditional distribution~$p(y|\boldsymbol{x})$, our proposal is to learn the joint distribution~$p(\boldsymbol{x},y)$, factorized as $p(\boldsymbol{x}|y)p(y)$, and to perform classification using Bayes' rule. As a proof-of-principle, here we implement this strategy by training a variational autoencoder for each class to be learned and by using importance sampling to estimate the likelihoods~$p(\boldsymbol{x}|y)$. This simple approach performs very well on a diverse set of continual learning benchmarks, outperforming generative replay and other existing baselines that do not store data.
\end{abstract}


\section{Introduction}
\label{sec:introduction}

Deep neural networks excel in supervised learning tasks, but only when all the classes to be learned are available at the same time. Incrementally training a deep neural network to distinguish between a gradually growing number of classes has turned out to be very challenging~\cite{rebuffi2017icarl,shin2017continual,vandeven2020brain,de2020continual}. Successful strategies for class-incremental learning generally either rely on storing a subset of the past data and/or on replaying (representations of) past data, both of which have important disadvantages. Storing data is not always possible in practice (\eg~due to safety/privacy concerns or because of limited storage capacity), while replay --- or rehearsal --- is computationally expensive as it involves constant retraining on past data.

These drawbacks have sparked recent interest in `rehearsal-free' continual learning~\cite{lomonaco2020rehearsal}, in which storing data or using replay are not allowed. In the past few years several methods have been proposed that can do class-incremental learning without replay or stored data~\cite{lomonaco2017core50,maltoni2019continuous,belouadah2020initialbmvc,hayes2020lifelong}. However, those methods rely on protocols with explicit task boundaries and/or their performance critically depends on the availability of a suitably pre-trained feature extractor.

In this paper, we put forward generative classification as a promising new strategy for class-incremental learning. Specifically, instead of training neural networks to directly learn the conditional distribution~$p(y|\boldsymbol{x})$, we propose to train them to learn the joint distribution~$p(\boldsymbol{x},y)$, factorized as $p(\boldsymbol{x}|y)p(y)$, and then to perform classification using Bayes' rule. A key benefit of this strategy is that it rephrases a challenging class-incremental learning problem as a more easily addressable task-incremental learning problem (see Section~\ref{sec:framework}).

To demonstrate the potential of generative classification for class-incremental learning, as a proof-of-principle we implement this strategy by training a variational autoencoder model for each class to be learned and by using importance sampling to estimate the class-conditional likelihoods during inference.
We find that such a straight-forward implementation of a generative classifier performs very well on a diverse range of class-incremental learning problems, outperforming generative replay and existing rehearsal-free methods. Moreover, this approach does not use replay, it does not store data, it can be applied to arbitrary class-incremental data streams (\ie~no need for task boundaries) and it does not rely on pre-trained networks, although if available those can be used effectively.


\section{Problem formulation}

In continual or incremental learning, an algorithm does not have access to all data at the same time, but it encounters the data in a sequence~\cite{parisi2019continual,delange2021continual,hadsell2020embracing}. Recently, three different types, or `scenarios', of continual learning have been described~\cite{vandeven2018three}: in task-incremental learning an algorithm must incrementally learn a set of clearly distinct tasks, in domain-incremental learning an algorithm must learn the same task but with changing contexts, and in class-incremental learning an algorithm must incrementally learn to distinguish between a growing number of classes. In this paper, we focus on class-incremental learning, which is generally considered to be the most challenging continual learning scenario~\cite{masana2020class,belouadah2020comprehensive,prabhu2020gdumb}.

\subsection{Class-incremental learning}
There are various different ways in which a class-incremental learning problem can be set up. This makes direct comparisons between studies challenging, even when they use the same datasets. We therefore start by discussing some important assumptions that vary between studies.

\subsubsection{Task-based \emph{vs.} task-free}
The goal of class-incremental learning is to learn, given a dataset $\mathcal{D} = \{x_i, y_i\}_{i = 1}^{n}$, a classification rule that maps an input $x\in\mathcal{X}$ to a predicted label $y\in\mathcal{Y}$. However, unlike in classical machine learning, the algorithm that must learn this mapping is not given access to the entire dataset at once. Instead, the data is made available according to a particular class-incremental protocol.

\begin{figure}[t]
\vskip -0.0in
\begin{center}
\includegraphics[width=0.99\columnwidth]{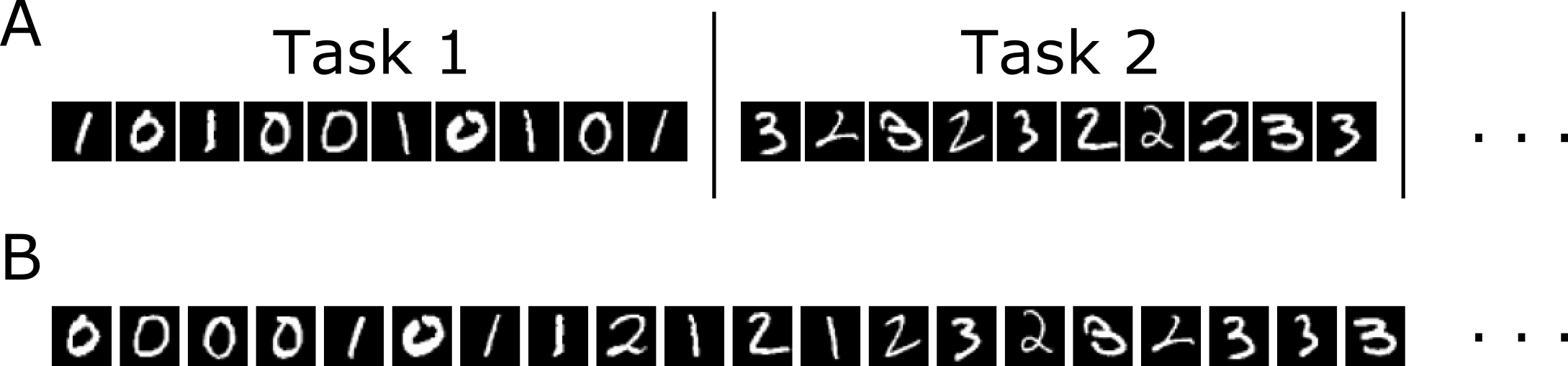}
\vskip 0.08in
\caption{Schematic illustrating the distinction between (A) task-based and (B) task-free class-incremental learning.}
\label{fig:learning_scenario}
\end{center}
\vskip -0.175in
\end{figure}

\vspace{-0.105in}
\paragraph{Task-based class-incremental learning}
A commonly used class-incremental learning protocol is to split up the dataset into distinct `tasks' (or `episodes'), whereby each task contains a different subset of classes [\eg~\citealp{shin2017continual,rebuffi2017icarl,vandeven2018three}]. The algorithm is then sequentially given access to the data of each task (Figure~\ref{fig:learning_scenario}A). Importantly, after transitioning from one task to the next, the data from the previous task is no longer available. During each task, the training data of that task could either be given to the algorithm all at once, or it might be presented according to a fixed stream that is not controlled by the algorithm (see Appendix~\ref{sec:assumptions}).

\vspace{-0.105in}
\paragraph{Task-free class-incremental learning}
\label{sec:task_free}
It has been argued that task-based protocols are not representative of real-world problems, and that the community should shift its focus to `task-free' continual learning~\cite{aljundi2019online,aljundi2019task,zeno2019task,hayes2020lifelong}. In a task-free protocol, the algorithm is presented with an arbitrary stream of data, without any prior knowledge about the structure of this stream (Figure~\ref{fig:learning_scenario}B). Many existing methods for class-incremental learning cannot deal with this setting, because they rely on the presence of `task boundaries' (see Table~\ref{tab:assumptions} in the Appendix for an overview). 

In general, benchmarks for task-free class-incremental learning need to include a protocol for how the data stream should be generated (\ie~they should specify when samples from each class are presented). An open, largely unaddressed research question relates to the development of a principled way to design such data streams. In this paper we side-step this question, because for the particular implementation of generative classifier considered here --- with a separate generative model for each class --- the actual class-incremental sequence of the data stream does not matter.

Task-free continual learning has also been referred to as `streaming' or `online' continual learning. In that case, sometimes additional constraints are that each training sample should only be presented once and that the mini-batch size should be one~\cite{aljundi2019online,hayes2020lifelong}. However, it is worth pointing out that these constraints relate to the sample efficiency of an algorithm and its robustness to noisy updates and, although they are topics worth studying, these are independent from the distinction between task-based and task-free class-incremental learning. For one of the benchmarks reported in this paper, we follow this more strict definition of streaming learning.

\subsubsection{Other critical assumptions}

\vspace{-0.03in}
\paragraph{Data storage}
An important assumption made by many class-incremental learning methods is that it is acceptable to store a limited amount of past samples in a memory buffer. The size of this memory buffer is typically one of the most important determinants of a method's performance~\cite{prabhu2020gdumb,balaji2020effectiveness}. In practice, storing data is not always possible (\eg~safety or privacy concerns), and in this study we do not allow data storage, a setting which has been referred to as memoryless class-incremental learning~\cite{belouadah2020initialbmvc}.

\vspace{-0.105in}
\paragraph{Pre-training}
Another assumption commonly made in the class-incremental learning literature, especially by studies that do not allow storing data, is that a suitably pre-trained network or feature extractor is available or that there is an extended, non-incremental initialization phase that can be used for pre-training [\eg~\citealp{maltoni2019continuous,vandeven2020brain,hayes2020lifelong,liu2020generative}]. While the importance of the assumption about data storage seems to be widely acknowledged, this assumption about pre-training has received less attention. Here we investigate the importance of pre-training by considering both benchmarks with pre-trained networks available (CIFAR-100 and CORe50) and benchmarks without (MNIST and CIFAR-10).


\section{Existing class-incremental learning methods}
\label{sec:background}

\subsection{Methods relying on stored data}
Many class-incremental learning methods store a subset of past data in a memory buffer. That data could be replayed when training on new data~\cite{lopez2017gradient,rolnick2019experienceneurips,chaudhry2019tiny}, they could be used as exemplars or prototypes to guide classification decisions~\cite{rebuffi2017icarl,de2020continual} or they could be used in other ways~\cite{wu2019large,hou2019learning,belouadah2020scail}. Important questions when storing data are which samples to store~\cite{pan2020continualneurips,mundt2020wholistic} and in what format~\cite{caccia2020online,hayes2020remind}. As discussed, we do not consider methods that store data.

\subsection{Generative replay}
If it is not possible to store data, an alternative is to replay generated `pseudo-data'~\cite{robins1995catastrophic}. This strategy has been shown to be successful for toy problems with relatively simple inputs~\cite{shin2017continual,vandeven2018three}, but it struggles on problems with more complex inputs, such as natural images~\cite{lesort2019generative,aljundi2019online}. Some recent studies have shown competitive performance with generative replay on class-incremental learning problems with natural images~\cite{vandeven2020brain,cong2020ganneurips,liu2020generative}, but the approaches in those studies depend on pre-trained networks (or on an extensive, non-incremental initialization phase~\cite{liu2020generative}).

We include two generative replay methods in our comparison: deep generative replay [DGR;~\citealp{shin2017continual}], which replays pixel-level representations, and brain-inspired replay [\mbox{BI-R};~\citealp{vandeven2020brain}], which replays latent feature representations.

\subsection{Regularization-based methods}
A popular strategy for continual learning is parameter regularization, which aims to minimize changes to parameters important for previously learned tasks. Examples of this strategy are elastic weight consolidation [EWC;~\citealp{kirkpatrick2017overcoming}] and synaptic intelligence [SI;~\citealp{zenke2017improved}]. Although it is well-established that these parameter regularization methods by themselves do not perform well in the class-incremental learning scenario~\cite{van2018generative,farquhar2018towards,hsu2018re}, we include them in our comparison for completeness. Some regularization-based methods can be interpreted as performing approximate Bayesian inference on the parameters of the neural network~\cite{kirkpatrick2017overcoming,nguyen2018variational,farquhar2019unifying} (\ie~Bayes' rule is used to find $p(\boldsymbol{\theta}|\mathcal{D})$, with $\mathcal{D}$ the observed data). Note that this is different from the generative classification strategy proposed in this paper, which uses Bayes' rule for the classification decision (\ie to find $p(y|\boldsymbol{x})$).

\subsection{Bias-correcting algorithms}
When a standard softmax-based classifier is trained on a class-incremental learning problem, it ends up predicting only the most recently seen classes~\cite{li2020energy}. It has been argued that this is due to a bias in the output layer~\cite{wu2019large,belouadah2020comprehensive}, and several recent class-incremental learning methods aim to correct this bias by making the magnitude of the output weights of all classes comparable.
Examples of this strategy are `CopyWeights with Re-init' [CWR;~\citealp{lomonaco2017core50}] and its improved version CWR+~\cite{maltoni2019continuous}. A disadvantage of these two methods is that they freeze the parameters of all hidden layers after the first task, so representation learning is limited. To address this, the method AR1 was proposed~\cite{maltoni2019continuous}, which is similar to CWR+ except that it does not freeze the hidden layers but regularizes them using a modified version of SI.

There are also several bias-correcting algorithms that rely on stored data from previously seen classes [\eg~\citealp{wu2019large,belouadah2020scail}], but as discussed we do not consider those methods here.

Related to these bias-correction algorithms, a trick to prevent large differences in the magnitude of the output weights between tasks in the first place, is to always only train on the classes from the current task (\ie~only include the output units of classes from the current task in the softmax-normalization, see Appendix~\ref{sec:taskIL_training} for details). Zeno \etal~\cite{zeno2019task} called this the `labels trick'.  A limitation of this trick is that there is no attempt to train the network to distinguish between classes from different tasks.

\subsection{Other methods}
Incremental linear discriminant analysis~\cite{pang2005incremental,kim2011incremental} is a popular method in the data mining community that is suitable for class-incremental learning. Until recently, this method had largely been ignored in the continual learning community, likely because it can only learn a linear classifier. However, a recent study applied this method --- now referred to as streaming linear discriminant analysis [SLDA;~\citealp{hayes2020lifelong}] --- to the features extracted by a fixed, pre-trained deep neural network, which resulted in impressive performance on several class-incremental learning problems. The main disadvantage of SLDA is that it is not capable of representation learning, which means that its performance will likely heavily depend on the availability of suitably pre-trained networks. Here we test this: on the benchmarks in this paper for which no pre-trained networks are available, we apply SLDA directly on the input space.


\section{Proposed strategy: generative classification}
\label{sec:method}

\subsection{General framework \& intuition}
\label{sec:framework}
In deep learning, the typical approach to classification is to train a neural network to directly learn the conditional distribution $p(y|\boldsymbol{x})$ that we are interested in, for example by training a feed-forward classifier with a softmax output layer using cross-entropy loss.
When all classes are available at the same time, this approach indeed works very well.
In the incremental setting, however, this direct approach breaks down. A softmax classifier trained in the standard way heavily over fits to the most recently seen classes, a phenomenon referred to as catastrophic forgetting.
A reason for this catastrophic forgetting is that, based on the most recently seen data, the empirical version of $p(y|\boldsymbol{x})$ --- which the softmax classifier aims to learn --- is indeed heavily biased towards the most recent classes. So far, as reviewed in Section~\ref{sec:background}, the dominant approach in the continual learning field has been to try to find methods and tricks to alleviate catastrophic forgetting.

Here we propose a shift of gears. Breaking with the traditional deep learning approach of training classifiers discriminatively, we propose to tackle class-incremental learning with generative classifiers. Rather than training deep neural networks to directly learn the conditional distribution $p(y|\boldsymbol{x})$, we propose to train them to learn the joint distribution $p(\boldsymbol{x},y)$ --- factorized as $p(\boldsymbol{x}|y)p(y)$ --- and to use Bayes' rule for classification.
The key benefit of this proposed strategy is that, in a class-incremental learning setting, based on the most recently seen data the empirical version of $p(\boldsymbol{x}|y)$ should not have any particular bias. Only the empirical version of $p(y)$ is biased, but learning this distribution without catastrophic forgetting is typically straight-forward (\eg~the number of times each label is observed could be counted) or not needed (\eg~if it can be assumed that all labels have the same prior probability).

\vspace{-0.1in}
\paragraph{Class-incremental problem becomes task-incremental}
Another way to describe the benefit of the proposed generative classifier strategy is that it turns a challenging class-incremental learning problem into an easier task-incremental learning problem.
This is the case because learning $p(\boldsymbol{x}|y)$ can be interpreted as a task-incremental problem whereby each `task' consists of learning a class-conditional generative model for a specific label~$y$.
An~important advantage of task-incremental learning is that it is possible to train networks with task-specific components~[\eg~\citealp{masse2018alleviating,serra2018overcomingicml,vogelstein2020omnidirectional}], or even to use completely separate networks for each task to be learned. This last insight is used for our proof-of-principle implementation of a generative classifier with a separate generative model for every class. Note however that it should be possible to use other task-incremental learning techniques to enable parameter sharing between these models (see also the discussion).

\subsection{Implementation:\! VAEs \& importance sampling}
In this paper, to demonstrate the potential of the proposed generative classification strategy, we implement a generative classifier by training a variational autoencoder [VAE;~\citealp{kingma2013auto}] model for each class to be learned\footnote{Note that this setup could also be described as a single VAE model with class-specific masks whereby for each class a different, non-overlapping subset of parameters is unmasked.} and by using importance sampling to estimate the likelihoods~$p(\boldsymbol{x}|y)$. For $p(y)$ we use a uniform distribution over all possible classes, as all benchmarks have an approximately equal amount of samples per class. In general, $p(y)$ could be learned from the data as well, for example by counting the number of times each class is observed in the training data.

\begin{table*}[t]
   \caption{\label{tab:benchmarks}Overview of the benchmarks used in this paper. Each benchmark consists of an image dataset split up into a number of distinct tasks, with all tasks containing an equal number of classes. Such a task-based design is not needed for our generative classifier, but it is used to enable a comparison with other methods. Within each task, the training data is presented to the algorithm in a random, i.i.d stream, with the number of iterations per task and the mini-batch size being part of the benchmark. Another important aspect of each benchmark is whether pre-trained models are available. For all benchmarks considered in this paper, storing data is not allowed.}
  \vskip 0.0in
  \begin{center}
  \begin{small}
  \begin{tabular}{l@{\hskip 0.2in}cc@{\hskip 0.35in}ccc@{\hskip 0.35in}c}
    \toprule
    & \multicolumn{2}{c}{\textbf{Dataset Info~~~~~~~~~~~~~~~~}} & \multicolumn{3}{c}{\textbf{Data-Stream Parameters~~~~~~~~~~~~~~}} & \textbf{Pretrained} \\
    & Classes & Image-type & Tasks & Iterations & Batch size & \textbf{Models?}\\
    \midrule \midrule
    \textbf{MNIST} & 10 & 28x28, grey & 5 & 2000 & 128 & - \\
    \textbf{CIFAR-10} & 10 & 32x32, RGB & 5 & 5000 & 256 & - \\
    \textbf{CIFAR-100} & 100 & 32x32, RGB & 10 & 5000 & 256 & ConvLayers \\
    \textbf{CORe50} & 10 & 128x128, RGB & 5 & single pass & 1 & ResNet18 \\
    \bottomrule
  \end{tabular}
  \end{small}
  \end{center}
  \vskip -0.1in
\end{table*}

\subsubsection{Variational autoencoder}
To learn the distribution $p(\boldsymbol{x}|y)$, we train a VAE model for each class to be learned. For the experiments on MNIST and CIFAR-10, a completely separate VAE model is learned for every class, while for the experiments on CIFAR-100 and CORe50 the lower, pretrained layers are shared between all models (see Section~\ref{sec:latent}).

A VAE model consists of an encoder $q_{\boldsymbol{\phi}}$ that maps an input $\boldsymbol{x}$ to a posterior distribution $q_{\boldsymbol{\phi}}(\boldsymbol{z}|\boldsymbol{x})$ in latent space, a decoder $p_{\boldsymbol{\theta}}$ that maps a latent variable $\boldsymbol{z}$ back to a distribution $p_{\boldsymbol{\theta}}(\boldsymbol{x}|\boldsymbol{z})$ in the input space and a prior distribution $p_{\text{prior}}(\boldsymbol{z})$. For the VAE models used in this paper, these distributions are given by:
\begin{align}
    q_{\boldsymbol{\phi}}(\boldsymbol{z}|\boldsymbol{x}) &= \mathcal{N}\left(\boldsymbol{z}\left|\,\boldsymbol{\mu}_{\boldsymbol{\phi}}^{(\boldsymbol{x})}, {\boldsymbol{\sigma}_{\boldsymbol{\phi}}^{(\boldsymbol{x})}}^2I\right.\right) \\
    p_{\boldsymbol{\theta}}(\boldsymbol{x}|\boldsymbol{z}) &= \mathcal{N}\left(\boldsymbol{x}\left|\,\boldsymbol{\mu}_{\boldsymbol{\theta}}^{(\boldsymbol{z})}, I\right.\right) \\
    p_{\text{prior}}(\boldsymbol{z}) &= \mathcal{N}\left(\boldsymbol{z}\left|\,\boldsymbol{0},I\right.\right)
\end{align}
whereby $\boldsymbol{\mu}_{\boldsymbol{\phi}}^{(\boldsymbol{x})}$ and $\boldsymbol{\sigma}_{\boldsymbol{\phi}}^{(\boldsymbol{x})}$ are the outputs of the encoder network when $\boldsymbol{x}$ is fed in, and $\boldsymbol{\mu}_{\boldsymbol{\theta}}^{(\boldsymbol{z})}$ is the output of the decoder network when $\boldsymbol{z}$ is fed in. For both the encoder network and the decoder network, we use deep neural networks. See Appendix~\ref{sec:per_benchmark} for full details on the architectures that are used for the different benchmarks. Importantly, for each benchmark, the architecture of the VAE models is chosen so that the \emph{total} number of parameters of the generative classifier is similar to the number of parameters used by generative replay.

The VAE models are trained by optimizing a variational lower bound to the likelihood $p_{\boldsymbol{\theta}}(\boldsymbol{x})=\int p_{\boldsymbol{\theta}}(\boldsymbol{x},\boldsymbol{z})d\boldsymbol{z}=\int p_{\boldsymbol{\theta}}(\boldsymbol{x}|\boldsymbol{z})p_{\text{prior}}(\boldsymbol{z})d\boldsymbol{z}$. This lower bound, or ELBO, is given by:
\begin{equation}
\begin{split}
  \mathcal{L}&_{\text{ELBO}} \left(\boldsymbol{\theta},\boldsymbol{\phi};\boldsymbol{x}\right) = E_{q_{\boldsymbol{\phi}}(\boldsymbol{z}|\boldsymbol{x})} \left[ \log\frac{p_{\boldsymbol{\theta}}(\boldsymbol{x},\boldsymbol{z})}{q_{\boldsymbol{\phi}}(\boldsymbol{z}|\boldsymbol{x})}\right] \\ 
  & = E_{q_{\boldsymbol{\phi}}(\boldsymbol{z}|\boldsymbol{x})}[\log p_{\boldsymbol{\theta}}(\boldsymbol{x}|\boldsymbol{z})] -  D_{KL}(q_{\boldsymbol{\phi}}(\boldsymbol{z}|\boldsymbol{x})||p_{\text{prior}}(\boldsymbol{z})) 
  \label{eq:elbo}
\end{split}
\end{equation}
where $D_{KL}$ is the Kullback-Leibler divergence. Full details of the VAE training are given in Appendix~\ref{sec:technical_vae}.

\subsubsection{Importance sampling}
To estimate the likelihoods $p(\boldsymbol{x}|y)$, we use importance sampling~\cite{rezende2015variationalicml,burda2016importance}. This means that the likelihood of a test sample~$\boldsymbol{x}$ under the VAE model of class $y$ is estimated using:
\begin{equation}
    p(\boldsymbol{x}|y)=\frac{1}{S}\sum_{s=1}^{S}\frac{p_{\boldsymbol{\theta}_y}\!\left(\boldsymbol{x}|\boldsymbol{z}^{(s)}\right) p_{\text{prior}}\left(\boldsymbol{z}^{(s)}\right)}{q_{\boldsymbol{\phi}_y}\!\left(\boldsymbol{z}^{(s)}|\boldsymbol{x}\right)}
\end{equation}
whereby $\boldsymbol{\theta}_y$ and $\boldsymbol{\phi}_y$ are the parameters of the VAE model of class $y$, $S$ is the number of importance samples and $\boldsymbol{z}^{(s)}$ is the $s^{\text{th}}$ importance sample drawn from $q_{\boldsymbol{\phi}_y}(\boldsymbol{z}|\boldsymbol{x})$. For the results in Table~\ref{tab:results}, we use $S=10,000$ importance samples for each likelihood estimation. The effect of reducing the number of importance samples is explored in Table~\ref{tab:samples}.

Based on Bayes' rule: $p(y|\boldsymbol{x}) \propto p(\boldsymbol{x}|y)p(y)$, classification is then done using:
\begin{equation}
\label{eq:inference}
    \hat{y}^{(\boldsymbol{x})} = \argmax_{y\in\mathcal{Y}} p(\boldsymbol{x}|y)p(y) = \argmax_{y\in\mathcal{Y}} p(\boldsymbol{x}|y)
\end{equation}
whereby $\hat{y}^{(\boldsymbol{x})}$ is the class label predicted by the generative classifier for test sample $\boldsymbol{x}$. Note that the last equality in Eq.~\ref{eq:inference} holds because, in this paper, $p(y)$ is modelled with a uniform distribution over all possible classes.

\subsection{When pre-trained models are available: \\reconstruction loss in the feature space}
\label{sec:latent}
The generative classifier approach described so far does not depend on the availability of pre-trained networks, as it is possible to train the full generative models from scratch. If pre-trained models are available, however, there are various ways in which they could be used. For example, suppose that pre-trained convolutional layers are available. One option would be to use these to initialize the convolutional layers of the encoder networks of the VAE models, and then to proceed with training in the standard way. Another option, which is the approach taken in this paper, is to use the pre-trained convolutional layers as a fixed feature extractor, and then to train the VAE models on the extracted features rather than on the raw inputs. An advantage of this second approach, which is reminiscent of recent studies that performed generative replay in the feature space~\cite{vandeven2020brain,liu2020generative}, is that it appears to be easier to learn good generative models for such extracted features, presumably because they are less complex than the raw inputs.


\section{Experiments}

In this section we test the above implementation of the proposed generative classification strategy on a diverse set of class-incremental learning benchmarks. On each benchmark, we compare our generative classifier with the applicable methods discussed in Section~\ref{sec:background} that do not store data (see Appendix~\ref{sec:technical_methods} for technical details of all compared methods). As far as possible, we use the same ``base network'' architecture and the same training settings for all compared methods.  Full details of the architectures and training settings used for each benchmark are provided in Appendix~\ref{sec:per_benchmark}.
Documented code for all experiments (including for all compared methods) is available online: \url{https://github.com/GMvandeVen/class-incremental-learning}.

\begin{table*}[h]
  \caption{\label{tab:results}Final test accuracy (as \%) of all compared methods on the different benchmarks. Evaluation is according to the ``class-incremental learning scenario'' or the ``single-headed setting'' (\ie~the model has to chose between all classes). Only methods that do not store data are included. All experiments were performed 10 times with different random seeds, reported are the means ($\pm$ SEMs) over these runs.}
  \vskip 0.0in
  \begin{center}
  \begin{small}
  \begin{tabular}{llp{1.9cm}p{1.9cm}p{1.9cm}p{1.9cm}}
    \toprule
    \textbf{Strategy} & \textbf{Method} & \textbf{~~MNIST} & \textbf{~~CIFAR-10} & \textbf{~~CIFAR-100} & \textbf{~~CORe50} \\
    \midrule \midrule
    \multirow{2}{*}{\it Baselines} & \it None & \it 19.92 ($\pm$ 0.02) & \it 18.74 ($\pm$ 0.29) & \it ~~7.96 ($\pm$ 0.11) & \it 18.65 ($\pm$ 0.26) \\
    & \it Joint & \it 98.23 ($\pm$ 0.04) & \it 82.07 ($\pm$ 0.15) & \it 54.08 ($\pm$ 0.27) & \it 71.85 ($\pm$ 0.30) \\
    \midrule
    \multirow{3}{*}{Generative Replay} & DGR & 91.30 ($\pm$ 0.60) & 17.21 ($\pm$ 1.88) & ~~9.22 ($\pm$ 0.24) & \,\,\,\,\,~~~~~~- \\
    & BI-R & \,\,\,\,\,~~~~~~- & \,\,\,\,\,~~~~~~- & 21.51 ($\pm$ 0.25) & 60.40 ($\pm$ 1.04) \\
    & BI-R\,+\,SI & \,\,\,\,\,~~~~~~- & \,\,\,\,\,~~~~~~- & 34.38 ($\pm$ 0.21) & 62.68 ($\pm$ 0.72) \\
    \midrule
    \multirow{2}{*}{Regularization} & EWC & 19.95 ($\pm$ 0.05) & 18.63 ($\pm$ 0.29)& ~~8.47 ($\pm$ 0.09) & 18.56 ($\pm$ 0.31) \\
     & SI  & 19.95 ($\pm$ 0.11) & 18.14 ($\pm$ 0.36) & ~~8.43 ($\pm$ 0.08) & 18.69 ($\pm$ 0.26) \\
    \midrule
    \multirow{4}{*}{Bias-correction} 
     & CWR & 32.48 ($\pm$ 2.64) & 18.37 ($\pm$ 1.61) & 21.90 ($\pm$ 0.68) & 40.28 ($\pm$ 1.13)\\
     & CWR+ & 37.20 ($\pm$ 3.11) & 22.32 ($\pm$ 1.08) & ~~9.34 ($\pm$ 0.25) & 40.12 ($\pm$ 1.06)\\
     & AR1 & 48.84 ($\pm$ 2.55) & 24.44 ($\pm$ 1.08) & 20.62 ($\pm$ 0.45) & 45.27 ($\pm$ 1.02) \\
     & Labels Trick & 32.46 ($\pm$ 1.95) & 18.43 ($\pm$ 1.31) & 23.68 ($\pm$ 0.26) & 42.59 ($\pm$ 1.03) \\
    \midrule
    Other & SLDA & 87.30 ($\pm$ 0.02) & 38.35 ($\pm$ 0.03) & 44.49 ($\pm$ 0.00) & \bf 70.80 ($\pm$ 0.00) \\
    \midrule
    \multicolumn{2}{l}{Generative Classifier} & \bf 93.79 ($\pm$ 0.08) & \bf 56.03 ($\pm$ 0.04) & \bf 49.55 ($\pm$ 0.06) & \bf 70.81 ($\pm$ 0.11)\\
    \bottomrule
  \end{tabular}
  \end{small}
  \end{center}
  \vskip -0.1in
\end{table*}

\subsection{Benchmarks}
\label{sec:benchmarks}

An overview of the benchmarks used in this paper is provided in Table~\ref{tab:benchmarks}. All benchmarks are set up as task-based, in order to be able to compare with current state-of-the-art class-incremental learning methods, even though our generative classifier can be applied to task-free protocols as well.\footnote{For the specific implementation of the generative classifier used in this paper, with a separate model for each class, the performance does not depend on the specific class-incremental sequence at all. The reason is that the class-specific VAE models are trained only on samples of their own class, and it therefore does not matter if those classes are intermingled in certain ways.} Important aspects of each benchmark are the number of tasks, the number of iterations per task, the mini-batch size and whether pre-trained models are available. For all benchmarks, within tasks the training data is always fed to the network in an i.i.d. stream, although some of the compared methods (EWC, and SLDA for the first task) additionally assume they can access a task's training data in one large batch (see Appendix~\ref{sec:assumptions}).

\subsubsection{MNIST}
The first benchmark is based on the MNIST dataset~\cite{lecun1998gradient}, which is split up into 5 tasks with 2 digits each. Following previous studies~\cite{van2018generative,hsu2018re}, this benchmark has 2000 iterations per task and a mini-batch size of 128. The base network for this benchmark is a fully-connected network with 2 hidden layers of 400 ReLU units and a softmax output layer. No pre-training is used.

\subsubsection{CIFAR-10 without pre-training}
For this benchmark the CIFAR-10 dataset~\cite{krizhevsky2009learning} is split up into 5 tasks with 2 classes each. The number of iterations per task for this benchmark is 5000 and the mini-batch size is 256. Following previous studies~\cite{lopez2017gradient,aljundi2019online,de2020continual}, the base network is a small version of ResNet18~\cite{he2016identity} with three times less feature maps across all layers. No pre-training is used.

\subsubsection{CIFAR-100 with pre-training on CIFAR-10}
This benchmark is taken from the study that proposed \mbox{BI-R}~\cite{vandeven2020brain}. The CIFAR-100 dataset~\cite{krizhevsky2009learning} is split up into 10 tasks with 10 classes each. There are 5000 iterations per task with mini-batch size of 256. The base network is a convolutional neural network with 5 pre-trained convolutional layers followed by 2 randomly initialized fully-connected layers with 2000 ReLU units and a softmax output layer. The convolutional layers were pre-trained on CIFAR-10. To enable a direct comparison, we use the exact same pre-trained convolutional layers as in~\cite{vandeven2020brain}, which were made publicly available by the authors.

\subsubsection{CORe50 with pre-training on ImageNet}
The final benchmark is based on the CORe50 dataset~\cite{lomonaco2017core50}. This dataset is made up of image-frames cropped from short 15 second videos of moving objects. There are 10 different classes, with each class represented in the dataset by 5 different objects that were each filmed in 11 different environments. As in~\cite{lomonaco2017core50,douillardlesort2021continuum}, we use the images from eight of these environments for training and the others for testing. This results in approximately $10,500$ training images per class. The dataset is split up into 5 tasks with 2 classes each. This benchmark follows the more strict definition of streaming learning: each training image is presented by itself (\ie~mini-batch size of 1) and only once. Following~\cite{hayes2020lifelong}, a standard ResNet18 pretrained on ImageNet is used as a fixed feature extractor. The base network on top of this feature extractor consists of one fully connected layer with 1024 ReLU units and a softmax output layer.

\begin{figure*}[h]
\vskip 0.0in
\begin{center}
\includegraphics[width=0.70\textwidth]{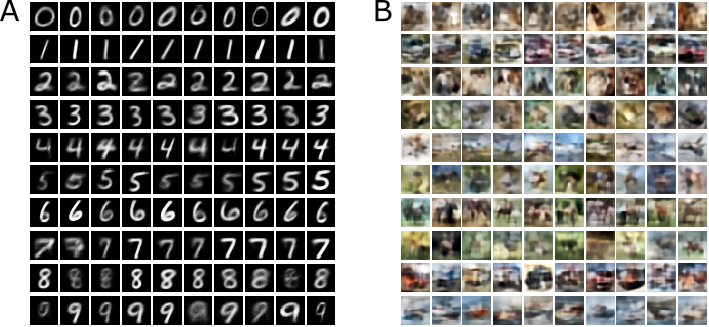}
\caption{Samples randomly drawn from the VAE models of the generative classifier for (A) MNIST and (B) CIFAR-10.}
\label{fig:samples}
\end{center}
\vskip 0.0in
\end{figure*}

\begin{table*}[h]
  \caption{\label{tab:replay}Comparison of the performance of the generative classifier with the performance of a softmax-based classifier discriminatively trained on samples from the VAE models of the generative classifier. Shown is the test accuracy (as \%) over all classes. All experiments were performed 10 times with different random seeds, reported are the means ($\pm$ SEMs) over these runs.}
  \vskip 0.0in
  \begin{center}
  \begin{small}
  \begin{tabular}{lp{1.9cm}p{1.9cm}p{1.9cm}p{1.9cm}}
    \toprule
     & \textbf{MNIST} & \textbf{CIFAR-10} & \textbf{CIFAR-100} & \textbf{CORe50} \\
    \midrule \midrule
    Generative classifier & \bf 93.79 ($\pm$ 0.08) & \bf 56.03 ($\pm$ 0.04) & \bf 49.55 ($\pm$ 0.06) & \bf 70.81 ($\pm$ 0.11)\\
    Discriminative classifier trained on generated samples & 85.93 ($\pm$ 0.43) & 13.71 ($\pm$ 0.61) & 33.84 ($\pm$ 0.14) & 47.86 ($\pm$ 1.77)\\
    \bottomrule
  \end{tabular}
  \end{small}
  \end{center}
  \vskip -0.1in
\end{table*}

\subsection{Results}
\label{sec:results}

Table~\ref{tab:results} shows the performance of our generative classifier on the four benchmarks described above, along with the performance of the methods discussed in Section~\ref{sec:background} that also do not store data. The generative classifier performed very strongly, comfortably outperforming all compared methods on three out of four benchmarks. Of special note are the substantial gaps with the generative replay variants, while these methods used similar number of parameters. Only on the CORe50 benchmark, in which an extensively pre-trained network was used and in which each sample was presented only once, the performance of the generative classifier was comparable to that of SLDA, while still substantially higher than that of the other compared methods.

An interesting result is that SLDA still performed competitively when it was applied directly on the raw inputs of MNIST and CIFAR-10. Although its performance was well below that of our generative classifier, it outperformed almost all other methods.

Another thing to note from these results is the modest performance of the bias-correction methods, especially on benchmarks where no pre-training was used. When pre-trained networks were available the relative performances of these methods improved, but they did not come close to those of the best performing methods.

\begin{table*}[h]
  \caption{\label{tab:samples}Performance of generative classifier as function of number of importance samples used for inference. Shown is test accuracy (as \%) over all classes. Experiments were performed 10 times with different random seeds, reported are the means ($\pm$ SEMs) over these runs.}
  \begin{center}
  \begin{small}
  \vskip 0.0in
  \begin{tabular}{lp{1.9cm}p{1.9cm}p{1.9cm}p{1.9cm}p{1.9cm}}
    \toprule
     & $S=1$ & $S=10$ & $S=100$ & $S=1,000$ & $S=10,000$ \\
    \midrule \midrule
    \textbf{MNIST} & 91.14 ($\pm$ 0.08) & 92.46 ($\pm$ 0.09) & 93.25 ($\pm$ 0.09) & 93.62 ($\pm$ 0.10) & 93.79 ($\pm$ 0.08) \\
    \textbf{CIFAR-10} & 50.86 ($\pm$ 0.10) & 54.64 ($\pm$ 0.09) & 55.43 ($\pm$ 0.10) & 55.83 ($\pm$ 0.09) & 56.03 ($\pm$ 0.04)\\
    \textbf{CIFAR-100} & 45.02 ($\pm$ 0.10) & 48.45 ($\pm$ 0.10) & 49.26 ($\pm$ 0.10) & 49.48 ($\pm$ 0.08) & 49.55 ($\pm$ 0.06)\\
    \textbf{CORe50} & 61.00 ($\pm$ 0.19) & 69.09 ($\pm$ 0.14) & 70.33 ($\pm$ 0.14) & 70.62 ($\pm$ 0.14) & 70.81 ($\pm$ 0.11)\\
    \bottomrule
  \end{tabular}
  \end{small}
  \end{center}
  \vskip -0.1in
\end{table*}

\subsection{Generative classification \emph{vs.} generative replay}
\label{sec:GRvsGC}
An intriguing result from the above comparisons is that our generative classifier consistently and sometimes substantially outperformed generative replay. This suggests that directly using generative models to perform classification might be a better strategy than using those models indirectly to generate replay for discriminatively training a classifier. However, it could be argued that this conclusion is not completely warranted by these results, as both strategies did not use the exact same generative models (even though the total number of parameters was similar). For generative replay one large generative model was incrementally trained on all classes, while for the generative classifier a series of smaller, separate generative models was trained.

To more directly compare generative classification and generative replay, we trained --- in an i.i.d.\!\!\! manner --- a softmax-based classifier on samples generated by the VAE models of the generative classifier (see Appendix~\ref{sec:offline_replay} for full details on this experiment). Another way to phrase this experiment is that a discriminative classifier was trained exclusively with `generative replay' produced by the same generative models as used by the generative classifier.
For all benchmarks, we found that the generative classifier substantially outperformed the discriminative classifier that was trained on its own samples (Table~\ref{tab:replay}). This suggests that, also when the same generative models are used, generative classification outperforms generative replay.

The results in Table~\ref{tab:replay} also indicate that the quality of the samples produced by the VAE models of our generative classifiers was not so good. To check this, we visualized samples drawn from the VAE models of the generative classifier for the MNIST and CIFAR-10 benchmarks (Figure~\ref{fig:samples}). While for MNIST the generated samples look reasonable, for CIFAR-10 they are indeed not great. This thus indicates that competitive class-incremental learning performance could be obtained by a generative classifier even without high-quality generative models.


\section{Discussion}
\label{sec:discussion}

Class-incremental learning is a challenging problem. So far the deep learning community has tackled this problem by directly learning a discriminative classifier, which only seems to work in combination with tricks such as pre-training, storing data or replay. Here we proposed an alternative strategy --- to learn a generative classifier --- and we showed that it can outperform generative replay and existing rehearsal-free methods.

An interesting finding from our comparison of class-incremental learning methods was the strong performance of SLDA~\cite{hayes2020lifelong}. It outperformed generative replay variants on three out of four benchmarks, and it achieved competitive performance even when applied directly on the raw inputs. We believe this strong performance can be explained because SLDA can be interpreted as a generative classifier. SLDA learns a mean vector $\boldsymbol{\mu}_y$ for each class $y$ and a covariance matrix $\Sigma$ that is shared between all classes. The generative model that SLDA implicitly assumes for each class $y$ is given by $p(\boldsymbol{x}|y)=\mathcal{N}\left(\boldsymbol{x}\left|\,\boldsymbol{\mu}_y,\Sigma\right.\right)$. SLDA is however ``a generative classifier in disguise'' because it does not explicitly compute the likelihoods during inference, since with its assumptions the decision boundaries implied by the underlying generative models can be computed analytically.

The main disadvantage of SLDA is that it can only learn linear classifiers. To further improve upon SLDA, it seems necessary to find a way to do representation learning in a class-incremental way.
This is exactly what our deep generative classifiers are able to do. Learning good representations is not easy, and it is not surprising that this ability comes at a cost of increased sample complexity.
However, (complex) representation learning is not a necessary component of our proposed strategy. When the amount of training data is small, or when the representations provided by a pre-trained network are already good, it is probably better to learn relatively simple generative models. 
Indeed, SLDA's performance can be seen as the minimal attainable performance for a generative classifier, upon which can be improved when sufficient data is available.


Compared with generative replay, an important advantage of generative classifiers is that training is less costly, as replay is not necessary. On the flip-side, inference (\ie~making a classification decision) with generative classifiers is relatively costly, as it involves computing/estimating the likelihood of a test sample under the generative model of each possible class. For our specific implementation, this seems especially problematic because a large number of importance samples tends to be needed for high precision likelihood estimates with VAE models~\cite{theis2016note}. 
For the results reported in Table~\ref{tab:results}, we used $10,000$ importance samples for each likelihood estimation. However, we found that the number of importance samples could be lowered substantially without large drops in performance (Table~\ref{tab:samples}). Even using just a single importance sample resulted in state-of-the-art class-incremental learning performance on three out of four benchmarks. Moreover, there are also other tricks that could speed up inference: it might be possible to use uncertainty estimates (which can be obtained from the generative models~\cite{mundt2020wholistic}) to inform the number of importance samples to use, or the classification decision could be made hierarchical (\eg~first decide whether it is a cat or a dog, then decide on the specific breed).

Another disadvantage of our specific implementation of the generative classifier is that a completely new generative model is learned for each new class. It could be questioned how scalable this is. In this regard, we believe it is important to point out three things. Firstly, to ensure a fair comparison between our generative classifier and generative replay, we controlled for the \emph{total} number of parameters. Secondly, as illustrated by SLDA, even using small or minimal generative models for each class can result in competitive performance. Finally, and perhaps most importantly, the main point of this paper is to highlight the potential of generative classification for class-incremental learning: our implementation with independent VAE models is a proof-of-principle. For practical applications, the generative models of the different classes should probably share substantial parts of their networks. Such sharing introduces the risk of interference, but it also opens up the possibility of positive transfer between the generative models. Importantly, as pointed out in Section~\ref{sec:framework}, learning the different class-conditional generative models is a task-incremental problem, which is an important simplification compared to the original class-incremental problem~\cite{vandeven2018three}. We therefore expect the question of how to optimally share parts of the generative models to be a fruitful topic for further research.


Finally, we highlight that VAE-based generative classifiers have also been shown to be able to perform strongly in terms of adversarial robustness~\cite{schott2019towards} and few-shot learning~\cite{mocanu2018one}, both or which are --- similar to continual learning --- hallmarks of human intelligence. Although those demonstrations were on relatively simple datasets, together with our results they tentatively suggest that a generative approach to classification might be a promising way to make classification with deep neural networks more human-like.

\subsection*{Acknowledgments}
\vskip -0.03in
We thank Siddharth Swaroop and Martin Mundt for useful comments. This research project has been supported by the Lifelong Learning Machines (L2M) program of the Defence Advanced Research Projects Agency (DARPA) via contract number HR0011-18-2-0025 and by the Intelligence Advanced Research Projects Activity (IARPA) via Department of Interior/Interior Business Center (DoI/IBC) contract number D16PC00003. Disclaimer: The views and conclusions contained herein are those of the authors and should not be interpreted as necessarily representing the official policies or endorsements, either expressed or implied, of DARPA, IARPA, DoI/IBC, or the U.S. Government.

{\small
\bibliographystyle{ieee_fullname}
\bibliography{references}
}

\newpage
~
\newpage

\appendix

\renewcommand\thefigure{\thesection.\arabic{figure}}   
\renewcommand\thetable{\thesection.\arabic{table}}   
\setcounter{figure}{0}  
\setcounter{table}{0}

\section{Experimental details}
\label{sec:details}

Documented PyTorch code to perform and build upon the experiments described in this paper is available online: \url{https://github.com/GMvandeVen/class-incremental-learning}.

\subsection{Technical details of the compared methods}
\label{sec:technical_methods}

This section provides the technical details of the methods compared against in Table~\ref{tab:results}. Most of the compared methods use deep neural networks that are trained by minimizing the multi-class cross-entropy loss (although some methods have additional terms in the loss function, see below) given by:
\begin{equation}
\mathcal{L}^{\text{CE}} \left(\boldsymbol{\theta};\boldsymbol{x},y\right) = -\log p_{\boldsymbol{\theta}}\left(Y=y|\boldsymbol{x}\right)
\label{eq:class_loss}
\end{equation}
whereby $p_{\boldsymbol{\theta}}\left(Y=y|\boldsymbol{x}\right)$ is the probability that input $\boldsymbol{x}$ belongs to class $y$ as predicted by the neural network with parameters $\boldsymbol{\theta}$. These probabilities are computed by performing a softmax normalization on the activations of the final output layer of the network. It is important to note that this softmax normalization is only performed over the output units of the classes that have been seen by the network up to that point in time. In other words, the networks are trained with an \emph{``expanding head''}~\cite{maltoni2019continuous} or only the classes seen so far are set as \emph{``active''}~\cite{vandeven2018three}. In this Appendix, this way of training a deep neural network is referred to as the ``standard way''.

In addition to the methods discussed below, the performance of the following two baselines is reported in Table~\ref{tab:results}:
\begin{description}
\item[- None:] The base neural network is sequentially trained on all tasks in the standard way. This baseline suffers from severe catastrophic forgetting, and is included as a lower bound.
\item[- Joint:] The base neural network is trained on all classes at the same time. For this baseline the same total number of iterations is used as for the incremental training protocols, with the difference that in each iteration the training data is randomly sampled from all classes rather than just from the classes in the current task. This baseline can be seen as an upper bound. 
\end{description}

\subsubsection{Deep generative replay}
With deep generative replay [DGR], following~\cite{shin2017continual}, a separate generative model is trained to generate input samples to be replayed. We use a variational autoencoder [VAE;~\citealp{kingma2013auto}] as generator. The encoder network of the VAE is always similar to the base network (except for the final softmax layer) and the decoder network is the mirror image of the encoder network; see Section~\ref{sec:per_benchmark} for the exact VAE architectures used for each benchmark. The classifier, or main model, is simply the base neural network.

Except on the first task, both the classifier and the generator are trained with replay. The replay is generated by sampling inputs from a copy of the generator, after which those inputs are labelled as the most likely class as predicted by a copy the classifier. The versions of the generator and classifier used to produce the replay are temporarily stored copies of both models after finishing training on the previous task.

Following~\cite{van2018generative}, for both the classifier and the generator, the total loss is a weighted sum of the loss on the data from the current task and the loss on the replayed data: $\mathcal{L} = \frac{1}{N_{\text{tasks so far}}} \mathcal{L}_{\text{current}} + (1-\frac{1}{N_{\text{tasks so far}}}) \mathcal{L}_{\text{replay}}$. For the main model, $\mathcal{L}_{\text{current}}$ and $\mathcal{L}_{\text{replay}}$ are the cross-entropy loss (see Eq.~\ref{eq:class_loss}). For the generator, $\mathcal{L}_{\text{current}}$ and $\mathcal{L}_{\text{replay}}$ are the VAE loss (see Eq.~\ref{eq:vae_loss}). In each iteration, the number of replayed samples is equal to the number of samples from the current task.


\subsubsection{Brain-inspired replay}
For brain-inspired replay [\mbox{BI-R}] we follow the exact protocol as described in~\cite{vandeven2020brain}, using the code released by the authors. We use all five of the proposed modifications (distillation, replay-through-feedback, conditional replay, gating based on internal context and internal replay). Because the internal replay component of \mbox{BI-R} relies on the availability of a pre-trained feature extractor, we do not use \mbox{BI-R} on the MNIST and CIFAR-10 benchmarks. Note that \mbox{BI-R} has a hyperparameter $X$, which controls the percentage of hidden units in the decoder that is masked per class (see Section~\ref{sec:gridsearch}).

\subsubsection{EWC \& SI}
For elastic weight consolidation [EWC;~\citealp{kirkpatrick2017overcoming}] and synaptic intelligence [SI;~\citealp{zenke2017improved}], the base neural network is trained in the standard way, except that a regularization term is added to the cross-entropy loss: $\mathcal{L} = \mathcal{L}^{\text{CE}} + \lambda \mathcal{L}^{\text{REG}}$, whereby hyperparameter $\lambda$ controls the regularization strength (see Section~\ref{sec:gridsearch}). This regularization term penalizes changes to parameters important for previously learned tasks.

\paragraph{EWC} The regularization term for EWC is given by:
\begin{equation}
  \mathcal{L}^{\text{REG}}\left(\boldsymbol{\theta}\right) = \sum_{k=1}^{K-1} \left( \frac{1}{2} \sum_{i=1}^{N_{\text{params}}} F_{ii}^{(k)} \left(\theta_i - \hat{\theta}_{i}^{(k)} \right)^2 \right)
  \label{eq:ewc}
\end{equation}
whereby $K$ is the current task, $\hat{\theta}_{i}^{(k)}$ are the parameters of the network after training on task $k$ and $F_{ii}^{(k)}$ is the estimated importance of parameter $i$ for task $k$. This last one is calculated as the $i^{\text{th}}$ diagonal element of the Fisher Information matrix of task $k$:
\begin{multline}
  F_{ii}^{(k)} = \\
  E_{\boldsymbol{x}\sim S^{(k)}} \left[ \sum_{c=1}^{N_{\text{classes}}} \tilde{y}_c^{(\textbf{x})} \left( \left. \frac{\delta\log p_{\boldsymbol{\theta}}\left(Y=c|\textbf{x}\right)}{\delta\theta_i} \right\rvert_{\boldsymbol{\hat{\theta}}^{(k)}} \right)^2\right]
  \label{eq:emp_fi}
\end{multline}
whereby $S^{(k)}$ is the training data of task $k$ and $\tilde{y}_c^{(\textbf{x})} = p_{\boldsymbol{\hat{\theta}}^{(k)}}\left(Y=c|\textbf{x}\right)$.

\paragraph{SI} The regularization term for SI is given by:
\begin{equation}
  \mathcal{L}^{\text{REG}}\left(\boldsymbol{\theta}\right) = \sum_{i=1}^{N_{\text{params}}} \Omega_{i}^{(K-1)} \left(\theta_i - \hat{\theta}_{i}^{(K-1)} \right)^2
  \label{eq:si}
\end{equation}
whereby $K$ is the current task, $\hat{\theta}_{i}^{(K-1)}$ are the parameters of the network after training on task $K-1$ and $\Omega_{i}^{(K-1)}$ is the estimated importance of parameter $i$ for the first $K-1$ tasks. To calculate these $\Omega_{i}^{(K-1)}$, after each task $k$, a per-parameter contribution to the change in loss is computed:
\begin{equation}
  \omega_i^{(k)} = \sum_{t=1}^{N^{k}_{\text{iters}}} \left(\hat{\theta}_i[t^{(k)}]-\hat{\theta}_i[\left(t-1\right)^{(k)}]\right) \frac{-\delta\mathcal{L}_{\text{total}}[t^{(k)}]}{\delta\theta_i}
  \label{eq:small_omega}
\end{equation}
whereby $N^{k}_{\text{iters}}$ is the number of iterations for task $k$, $\hat{\theta}_i[t^{(k)}]$ is the value of the $i^{\text{th}}$ parameter after the $t^{\text{th}}$ training iteration on task $k$ and $\frac{\delta\mathcal{L}_{\text{total}}[t^{(k)}]}{\delta\theta_i}$ is the gradient of the loss with respect to the $i^{\text{th}}$ parameter during the $t^{\text{th}}$ training iteration on task $k$. The $\Omega_{i}^{(K-1)}$ are then calculated as:
\begin{equation}
  \Omega_{i}^{(K-1)} = \sum_{k=1}^{K-1} \frac{\omega_i^{(k)}}{\left(\Delta_i^{(k)}\right)^2+\xi}
  \label{eq:omega}
\end{equation}
whereby $\xi$ is a dampening term that was set to 0.1 and $\Delta_i^{(k)} = \hat{\theta}_i[{N^k_{\text{iters}}}^{(k)}]-\hat{\theta}_i[0^{(k)}]$, where $\hat{\theta}_i[0^{(k)}]$ is the value of parameter $i$ when training on task $k$ started.

\subsubsection{CWR, CWR+ \& AR1}

\paragraph{CWR} For the method `CopyWeights with Re-init' [CWR;~\citealp{lomonaco2017core50}], the base neural network is trained on the first task in the standard way. After the first task, all parameters of the network are frozen except for the parameters of the output layer. For the parameters of the output layer, two copies are maintained: a temporary version denoted $\textbf{tw}$, and a `consolidated' version denoted $\textbf{cw}$. Training is done with $\textbf{tw}$. Before starting training on each task, $\textbf{tw}$ is randomly re-initialized. After finishing training on each task, the parameters in $\textbf{tw}$ corresponding to the classes of that task are copied over into $\textbf{cw}$. For testing, $\textbf{cw}$ is used.

\paragraph{CWR+} An improved version of CWR, called CWR+, was proposed in~\cite{maltoni2019continuous}. CWR+ has two differences compared to CWR. First, before each task, the parameters in $\textbf{tw}$ are set to zero rather than randomly \mbox{(re-)}initialized. Second, after each task, the parameters in $\textbf{tw}$ are first standardized by subtracting their mean (with the mean taken over all classes seen up to that point, and with a separate mean for the weights and the biases), and then the standardized parameters corresponding to the classes of that task are copied over into $\textbf{cw}$.

\paragraph{AR1} Building upon CWR+, Maltoni \& Lomonaco~\cite{maltoni2019continuous} also proposed AR1. This method is similar to CWR+, except that the parameters of the hidden layers are not frozen after the first task. Instead, a modified version of SI is used. The first modification is that SI is only used for the parameters of the hidden layers and not for the parameters of the output layer. The second modification is that $\Omega_{i}^{(K-1)}$ in Eq.~\ref{eq:si} is replaced by $\tilde{\Omega}_{i}^{(K-1)} = \max \left\{\Omega_{i}^{(K-1)}, \Omega_{\text{max}}\right\}$, with $\Omega_{\text{max}}$ a newly introduced hyperparameter that limits the extent to which each parameter could be regularized (see Section~\ref{sec:gridsearch}).

\subsubsection{Labels trick}
\label{sec:taskIL_training}
For the `labels trick'~\cite{zeno2019task}, the base network is trained in the standard way, except that always only the classes from the current task are set as `active' (see first paragraph of Section~\ref{sec:technical_methods}). This means that the softmax normalization is only performed over the output units of those classes, and that the network is therefore only trained on the classes from the current task. Another way to phrase this is that the network is trained as if it is trained on a task-incremental learning problem~\cite{vandeven2018three}. A fundamental limitation of this trick is that the network is never trained to learn to distinguish between classes from different tasks.

\subsubsection{SLDA}
The method streaming linear discriminant analysis [SLDA;~\citealp{hayes2020lifelong}] learns a linear classifier of the form:
\begin{equation}
    \hat{y}=\argmax_{c\in \mathcal{Y}} \left\{ \boldsymbol{w}_c^{\text{T}} \boldsymbol{x} + b_c \right\}
\end{equation}
whereby $\boldsymbol{w}_c$ is the $c^{\text{th}}$ row of weight matrix $\textbf{W}$, $b_c$ is the $c^{\text{th}}$ element of bias vector $\boldsymbol{b}$  and $\hat{y}$ is the predicted class label. 

To learn $\textbf{W}$ and $\boldsymbol{b}$, SLDA computes for each class $y$ a mean vector $\boldsymbol{\mu}_y$ and associated count $n_y$, as well as a single covariance matrix $\boldsymbol{\Sigma}$ that is shared between all classes. Updates to $\boldsymbol{\mu}_y$ and $n_y$ are done in a ``pure streaming'' manner: they are initialized at zero and for each new training sample $(\boldsymbol{x},y)$ that arrived at time $t$, they are updated as:
\begin{align}
    \boldsymbol{\mu}_y^{(t+1)} &= \frac{n_y^{(t)}\boldsymbol{\mu}_y^{(t)}+\boldsymbol{x}}{n_y^{(t)}+1} \\
    n_y^{(t+1)} &= n_y^{(t)}+1
\end{align}
with $\boldsymbol{\mu}_y^{(t)}$ and $n_y^{(t)}$ the versions of $\boldsymbol{\mu}_y$ and $n_y$ at time $t$. The covariance matrix is initialized on the first task using the Oracle Approximating Shrinkage estimator~\cite{chen2010shrinkage}, which is a batch-wise computation. On subsequent tasks, updates to $\boldsymbol{\Sigma}$ are done in a streaming manner: 
for each new training sample $(\boldsymbol{x},y)$ that arrives at time $t$, the following update is done:
\begin{equation}
    \boldsymbol{\Sigma}^{(t+1)} = \frac{t\boldsymbol{\Sigma}^{(t)}+\boldsymbol{\Delta}^{(t)}}{t+1}
\end{equation}
with $\boldsymbol{\Delta}^{(t)}= \frac{t}{t+1}\left(\boldsymbol{x}-\boldsymbol{\mu}_y^{(t)}\right)\left(\boldsymbol{x}-\boldsymbol{\mu}_y^{(t)}\right)^{\text{T}}$ and $\boldsymbol{\Sigma}^{(t)}$ the version of $\boldsymbol{\Sigma}$ at time $t$.

To perform classification, the rows of $\textbf{W}$ and the elements of $\boldsymbol{b}$ are then computed as:
\begin{align}
    \boldsymbol{w}_c &= \boldsymbol{\Lambda}\boldsymbol{\mu}_c \\
    b_c &= \boldsymbol{\mu}_c^{\text{T}}\boldsymbol{\Lambda}\boldsymbol{\mu}_c
\end{align}
where $\boldsymbol{\Lambda}=\left[(1-\epsilon)\boldsymbol{\Sigma+\epsilon\textbf{I}}\right]^{-1}$ and $\epsilon=0.0001$.

With SLDA it is not possible to train the parameters of a deep neural network. However, as pointed out in~\cite{hayes2020lifelong}, it is possible to use a pre-trained deep neural network as feature extractor. 
On the CIFAR-100 and CORe50 benchmarks, we use the pre-trained networks that are available for those benchmarks as feature extractor. On the MNIST and CIFAR-10 benchmarks, for which no pre-trained networks are available, we apply SLDA directly on the raw inputs.

\subsection{VAE training}
\label{sec:technical_vae}
VAE models are trained both for the generative classifier and for the generative replay variants. In both cases, the set-up of the VAE models and their training is similar, except that the VAE models of the generative classifier are smaller than those used for generative replay (see Section~\ref{sec:per_benchmark}).

Each VAE model consists of two deep neural networks: (1) an encoder network, parameterized by $\boldsymbol{\phi}$, mapping an input $\boldsymbol{x}$ to the mean $\boldsymbol{\mu}_{\boldsymbol{\phi}}^{(\boldsymbol{x})}$ and standard deviation $\boldsymbol{\sigma}_{\boldsymbol{\phi}}^{(\boldsymbol{x})}$ of the posterior distribution $q_{\boldsymbol{\phi}}(\boldsymbol{z}|\boldsymbol{x}) = \mathcal{N}\left(\boldsymbol{z}\left|\,\boldsymbol{\mu}_{\boldsymbol{\phi}}^{(\boldsymbol{x})}, {\boldsymbol{\sigma}_{\boldsymbol{\phi}}^{(\boldsymbol{x})}}^2I\right.\right)$ over the latent variables $\boldsymbol{z}$; and (2) a decoder network, parameterized by $\boldsymbol{\phi}$, mapping a latent variable vector $\boldsymbol{z}$ to a reconstructed input $\boldsymbol{\mu}_{\boldsymbol{\theta}}^{(\boldsymbol{z})}$, which is used as the mean of a Gaussian observer model $p_{\boldsymbol{\theta}}(\boldsymbol{x}|\boldsymbol{z}) = \mathcal{N}\left(\boldsymbol{x}\left|\,\boldsymbol{\mu}_{\boldsymbol{\theta}}^{(\boldsymbol{z})}, I\right.\right)$. The prior distribution over the latent variables $\boldsymbol{z}$ is the standard normal distribution: $p_{\text{prior}}(\boldsymbol{z}) = \mathcal{N}\left(\boldsymbol{z}\left|\,\boldsymbol{0},I\right.\right)$.


The parameters of these two networks are trained by maximizing a variational lower bound to the likelihood, or ELBO (see Eq.~\ref{eq:elbo} in the main text), which is equivalent to minimizing the following loss function:
\begin{equation}
\begin{split}
  \mathcal{L}&^{\text{VAE}} \left(\boldsymbol{\theta},\boldsymbol{\phi};\boldsymbol{x}\right) = E_{q_{\boldsymbol{\phi}}(\boldsymbol{z}|\boldsymbol{x})} \left[ -\log\frac{p_{\boldsymbol{\theta}}(\boldsymbol{x},\boldsymbol{z})}{q_{\boldsymbol{\phi}}(\boldsymbol{z}|\boldsymbol{x})}\right] \\ 
  & = E_{q_{\boldsymbol{\phi}}(\boldsymbol{z}|\boldsymbol{x})}[-\log p_{\boldsymbol{\theta}}(\boldsymbol{x}|\boldsymbol{z})] +  D_{KL}(q_{\boldsymbol{\phi}}(\boldsymbol{z}|\boldsymbol{x})||p_{\text{prior}}(\boldsymbol{z})) \\
  & = \mathcal{L}^{\text{recon}}\left(\boldsymbol{\theta},\boldsymbol{\phi};\boldsymbol{x}\right) + \mathcal{L}^{\text{latent}}\left(\boldsymbol{\phi};\boldsymbol{x}\right)
  \label{eq:vae_loss}
\end{split}
\end{equation}
whereby $D_{KL}$ is the Kullback-Leibler divergence. The first term in this loss function can be simplified to:
\begin{equation}
  \mathcal{L}^{\text{recon}}\left(\boldsymbol{\theta},\boldsymbol{\phi};\boldsymbol{x}\right) = E_{\boldsymbol{\epsilon}\sim\mathcal{N}(\textbf{0},\boldsymbol{I})} \left[ \sum_{i=1}^{N_{\text{inputs}}} \left(x_i - \mu_{\boldsymbol{\theta},i}^{\tilde{\boldsymbol{z}}}\right)^2 \right]
  \label{eq:vae_recon}
\end{equation}
whereby $x_i$ is the $i^{\text{th}}$ element of the original input $\boldsymbol{x}$ and $\mu_{\boldsymbol{\theta},i}^{\tilde{\boldsymbol{z}}}$ is the $i^{\text{th}}$ element of the decoded input $\boldsymbol{\mu}_{\boldsymbol{\theta}}^{\tilde{\boldsymbol{z}}}$, with $\tilde{\boldsymbol{z}}=\boldsymbol{\mu}_{\boldsymbol{\phi}}^{(\textbf{x})} + \boldsymbol{\sigma}_{\boldsymbol{\phi}}^{(\textbf{x})} \odot \boldsymbol{\epsilon}$. We estimate Eq.~\ref{eq:vae_recon} with a single sample of $\boldsymbol{\epsilon}$ for each datapoint.

The second term in Eq.~\ref{eq:vae_loss} is calculated analytically\footnote{See Appendix B in~\cite{kingma2013auto} for the full derivation.}:
\begin{equation}
  \mathcal{L}^{\text{latent}}(\boldsymbol{\phi};\boldsymbol{x}) = \frac{1}{2}\sum_{j=1}^{N_{\text{latent}}}\left(1+\log({\sigma_{\boldsymbol{\phi},j}^{(\textbf{x})}}^2)-{\mu_{\boldsymbol{\phi},j}^{(\textbf{x})}}^2-{\sigma_{\boldsymbol{\phi},j}^{(\textbf{x})}}^2\right)
  \label{eq:vae_latent}
\end{equation}
where $\mu_{\boldsymbol{\phi},j}^{(\textbf{x})}$ and ${\sigma_{\boldsymbol{\phi},j}^{(\textbf{x})}}$ are the $j^{\text{th}}$ elements of $\boldsymbol{\mu}_{\boldsymbol{\phi}}^{(\textbf{x})}$ and $\boldsymbol{\sigma}_{\boldsymbol{\phi}}^{(\textbf{x})}$, and $N_{\text{latent}}$ is the dimension of the latent space.

\subsection{Architectures \& training settings}
\label{sec:per_benchmark}

Neural network training is always done using the Adam-optimizer~\cite{kingma2014adam} with default settings (\ie~$\beta_1=0.9$, $\beta_2=0.999$). Depending on the benchmark, the learning rate is 0.001 (MNIST and CIFAR-10) or 0.0001 (CIFAR-100 and CORe50).

\subsubsection{MNIST}
For the MNIST benchmark, the base neural network has two fully-connected hidden layers with 400 ReLU units each, followed by a softmax output layer.

For DGR, the generative model is a symmetric VAE with both the encoder network and the decoder network similar to the base network (\ie~two fully-connected layers with 400 units each). 
The dimension of the latent space is 100. The same architecture was used in previous studies~\cite{vandeven2018three,hsu2018re}.

For our generative classifier implementation, we use VAE models with both the encoder network and the decoder network consisting of two fully-connected layers with 85 units. The dimension of the latent space is 5.

\subsubsection{CIFAR-10}
For the CIFAR-10 benchmark, following several previous studies~\cite{lopez2017gradient,aljundi2019online,de2020continual}, the base neural network is a slimmed down version of ResNet18~\cite{he2016identity}. In each layer, this version has approximately three times less channels than the the standard ResNet18: it has 20, 20, 40, 80 and 160 channels in the subsequent layers (instead of 64, 64, 128, 256 and 512). After the final residual block, global average pooling is applied, which is followed by the softmax output layer.

For DGR, the generative model is a VAE whereby the encoder network is similar to the base neural network (except that no pooling is used and there is no softmax output layer) and the decoder network is the mirror image of the encoder network. The dimension of the latent space is 100.

The VAE models that are used for the generative classifier have an encoder network that consists of three standard convolutional layers (with 15, 30 and 60 channels; each layer used batchnorm, ReLU non-linearities, a 3x3 kernel, a padding of 1 and a stride of 2), a decoder network that is the mirror image of the encoder network and a latent space of dimension 100.

\subsubsection{CIFAR-100}
For the CIFAR-100 benchmark, following~\cite{vandeven2020brain}, the base neural network has five pre-trained convolutional layers (16, 32, 64, 128 and 254 channels) followed by two randomly-initialized fully-connected layers with 2000 ReLU units and a softmax output layer. The convolutional layers are the same ones as in~\cite{vandeven2020brain}: they use batch-norm, ReLU non-linearities, a 3x3 kernel, a padding of 1, and a stride of 1 (first layer) or 2 (all other layers). They have been pre-trained on CIFAR-10 for 100 epochs using the ADAM-optimizer ($\beta_1=0.9$, $\beta_2=0.999$), a learning rate of 0.0001 and a mini-batch size of 256. On this benchmark, all methods are run twice: once with the pre-trained convolutional layers frozen and once with those layers plastic. Reported for each method in Table~\ref{tab:results} in the main text is the variant that performed best. For all methods this is the variant with the convolutional layers frozen, except for AR1 and the joint training baseline.


The generative model for DGR is a symmetric VAE with as encoder network the base neural network, as decoder network a mirror image of the encoder network and latent space dimension of 100. For \mbox{BI-R}, the combined classifier/generator model is the same as the VAE for DGR, except that the deconvolutional layers are removed from the decoder network and that a softmax output layer is appended to the top layer of the encoder network.

For the generative classifier, reminiscent of the approach of \mbox{BI-R}, we train the VAE models on the features extracted by the pre-trained convolutional layers rather than on the raw inputs. That means that the reconstruction loss of the VAE models is in the feature space instead of at the pixel level. The VAE models that we use have an encoder network and a decoder network both consisting of one fully-connected hidden layer with 85 ReLU units and a latent space with dimension 20.

\begin{table*}[h]
  \caption{\label{tab:hyperparams}Overview of the explored and selected hyperparameter values.}
  \vskip 0.15in
  \begin{center}
  \begin{small}
  \begin{tabular}{lcccccc}
    \toprule
    \multirow{2}{*}{\textbf{Method}} & \multirow{2}{*}{\textbf{Param}} & \multirow{2}{*}{\textbf{Explored range}} & \multicolumn{4}{c}{\textbf{Selected values}} \\
     & & & \textbf{MNIST} & \textbf{CIFAR-10} & \textbf{CIFAR-100} & \textbf{CORe50} \\
    \midrule \midrule
    BI-R & $X$ & $[0,10,20,...,80,90]$ & - & - & 70 & 0\\
    \midrule
    \multirow{2}{*}{BI-R + SI} & $X$ & $[0,20,40,60,80]$ & - & - & 60 & 60 \\
     & $\lambda$ &  $[0,0.001,0.01,...,10^{8},10^{9}]$ & - & - & $10^8$ & 0.01 \\
    \midrule
    SI & $\lambda$ & $[0,0.001,0.01,...,10^{8},10^{9}]$ & $10^3$ & 1 & 1 & 10\\
    \midrule
    EWC & $\lambda$ & $[0,0.1,1,...,10^{6},10^{7}]$ & $10^6$ & 10 & 100 & 10 \\
    \midrule
    \multirow{2}{*}{AR1} & $\lambda$ &  $[0,0.001,0.01,...,10^{8},10^{9}]$ & 10 & 100 & 100 & 1 \\
     & $\Omega_{\text{max}}$ & $[0.0001,0.001,...,10,100]$ & 0.01 & 0.1 & 10 & 0.1 \\
    \bottomrule
  \end{tabular}
  \end{small}
  \end{center}
  \vskip -0.1in
  \vskip 0.1in 
\end{table*}

\subsubsection{CORe50}
For the CORe50 benchmark, the base neural network is a standard ResNet18 that has been pre-trained on ImageNet (downloaded from PyTorch), followed by one fully-connected layer with 1024 ReLU units and a softmax output layer. For all methods, the parameters of the ResNet18 are frozen, and only the fully-connected layer and the output layer are trained.

On this benchmark we do not perform DGR, because we do not believe that training a pixel-level generative model in a pure streaming manner on CORe50 stands any chance of success (except perhaps if the full generative model has been pre-trained). For \mbox{BI-R}, the reconstruction objective is placed at the level of the features extracted by the pre-trained ResNet18. The model used for \mbox{BI-R} has an encoder network and a decoder network that both consist of one fully-connected hidden layer with 1024 ReLU units, it has a softmax output layer on top of the encoder network and the dimension of the latent space is 200.

The VAE models of the generative classifier are trained on the features extracted by the pre-trained ResNet18. These VAE models have no hidden layers (\ie~there is only a fully-connected layer from the ResNet embeddings to the latent space) and a latent space of dimension 110.

\subsection{Direct comparison between generative classification and generative replay}
\label{sec:offline_replay}
For the experiments described in Section~\ref{sec:GRvsGC} in the main text, a softmax-based classifier is trained in an i.i.d. manner on samples generated by the VAE models of the generative classifier. The classifier used for these experiments is the base neural network of each benchmark. This network is trained for the same number of iterations (and using the same mini-batch size and training settings) as for the joint training baseline. The only difference with the joint training baseline is that each mini-batch is made up of samples generated by the VAE models rather than by samples from the original training data. The samples are generated by first randomly sampling a class from all possible classes, after which a sample is drawn from the VAE model of that class.

\subsection{Hyperparameter searches}
\label{sec:gridsearch}
Several of the methods we compare against have one or more hyperparameters. Hyperparameters in continual learning can be problematic, because typically they are set by running a method on the full benchmark with a range of different hyperparameter-values. This means that these parameters are `learned' in a non-continual way, see also the discussion in the Appendix of~\cite{vandeven2018three}. Nevertheless, to give the methods we compare against the best chance, we select their hyperparameters based on gridsearches (see Table~\ref{tab:hyperparams}). These gridsearches are performed with a single random seed. The results in Table~\ref{tab:results} in the main text are then obtained with ten different random seeds.

\begin{table*}[h!]
 \renewcommand\thetable{B.1}
  \caption{\label{tab:assumptions}Overview of class-incremental learning variants to which the methods that are compared in this paper can be applied to. The symbols $^{(+)}$ and $^{(*)}$ indicate nuances that are discussed in Section~\ref{sec:assumptions}.}
  \vskip 0.15in
  \begin{center}
  \begin{small}
  \begin{tabular}{llccc}
    \toprule
    \multirow{2}{*}{\textbf{Strategy}} & \multirow{2}{*}{\textbf{Method}} & \emph{\textbf{Task-based}} & \emph{\textbf{Task-based}} & \emph{\textbf{Task-free}} \\
     & & \emph{\textbf{batch-wise}} & \emph{\textbf{streaming}} & \emph{\textbf{streaming}} \\
    \midrule \midrule
    \multirow{2}{*}{Generative Replay} & DGR & v & v & -\\
    & BI-R & v & v & -\\
    \midrule
    \multirow{2}{*}{Regularization} & EWC & v & - & -\\
     & SI & v & v & -\\
    \midrule
    \multirow{3}{*}{Bias-correction} & CWR / CWR+& v & v & v$^{(+)}$\\
     & AR1 & v & v & v$^{(+)}$\\
     & Labels Trick & v & v & -\\
    \midrule
    Other & SLDA & v & v$^{(*)}$ & v$^{(*)}$\\
    \midrule
    \multicolumn{2}{l}{Generative Classifier} & v & v & v\\   
    \bottomrule
  \end{tabular}
  \end{small}
  \end{center}
  \vskip -0.1in
  \vskip 0.15in 
\end{table*}

\section{A further distinction: batch-wise \emph{vs.} \mbox{streaming}}
\label{sec:assumptions}
\setcounter{figure}{0}  
\setcounter{table}{0}

Within task-based class-incremental learning, a further distinction can be made depending on whether within each task the algorithm is given free access to all data at once (``task-based batch-wise'') or whether the task's data is fed to the algorithm according to a fixed stream outside of the control of the algorithm (``task-based streaming'').
This distinction is important because some continual learning methods perform at each task boundary a consolidation operation that requires cycling over the training data of that task (\eg~estimating the Fisher Information matrix in EWC), so those methods are only suitable for task-based batch-wise learning.
Another difference is that with batch-wise learning, training settings such as number of iterations and mini-batch size can be decided on by the algorithm itself, while in streaming learning these are part of the benchmark.

Table~\ref{tab:assumptions} provides an overview of which methods can be applied in different class-incremental learning settings:
\begin{itemize}
\item The generative replay methods (DGR and \mbox{BI-R}) require task boundaries in order to decide when to update the copy of the models used to generate the replay. The current version of these methods are therefore not suitable for task-free continual learning. Neither DGR or \mbox{BI-R} makes assumptions about the way the data within each task is encountered, so both methods can be used for task-based streaming learning.
\item The regularization-based methods (EWC and SI) require task boundaries in order to decide when to update their regularization term, and the standard version of these methods are therefore not suitable for task-free continual learning (although see~\cite{zeno2019task} for a possible work around). EWC additionally makes the assumption that at each task boundary it is possible to make another pass over the training data of that task to estimate the Fisher Information matrix. EWC can therefore only be applied to task-based batch-wise learning, while SI can also be used for task-based streaming learning.
\item The bias correction methods (CWR, CWR+, AR1 and the labels trick) do not make any assumptions about how the data within each task is encountered, and all of them are applicable to both variants of task-based learning. The labels trick relies on specified tasks for the task-specific training, so this approach is not suitable for the task-free setting. The original versions of CWR, CWR+ and AR1 are also not compatible with task-free continual learning because they rely on the task boundaries for consolidating the (task-specific) weights of the output layer, but recently an adaption of these methods has been proposed to make them suitable for the task-free setting~\cite{lomonaco2020rehearsal}.
\item In principle, SLDA is a streaming method that is generally applicable to both task-based and task-free class-incremental learning. However, SLDA does make the assumption that its covariance matrix can be initialized on the first task (or in a separate `base initialization phase') in a batch-wise operation.
\item The generative classifier strategy proposed in this paper can be applied to both task-based and task-free class-incremental learning, as it does not make any assumptions about task boundaries or about being able to access larger amounts of data at the same time. In fact, for the specific implementation of the generative classifier that was considered in this paper --- with a separate VAE model for each class to be learned --- the specific sequence in which the different classes are presented does not matter at all, because the model for each class is trained only on data from its own class.
\end{itemize}

\end{document}